# Early Detection of Mental Health Issues Using Social Media Posts

Qasim Bin Saeed
Department of Civil Engineering
University of Manitoba
Winnipeg, Canada
binsaeeq@myumanitoba.ca

Ijaz Ahmed
Department of Mechanical Engineering
TU Bergakademie Freiberg
Freiberg, Germany
ijaz.ahmed@bwl.tu-freiberg.de

*Abstract*— **The increasing prevalence of mental health disorders, such as depression, anxiety, and bipolar disorder, calls for immediate need in developing tools for early detection and intervention. Social media platforms, like Reddit, represent a rich source of user-generated content, reflecting emotional and behavioral patterns. In this work, we propose a multi-modal deep learning framework that integrates linguistic and temporal features for early detection of mental health crises. Our approach is based on the method that utilizes a BiLSTM network both for text and temporal feature analysis, modeling sequential dependencies in a different manner, capturing contextual patterns quite well. This work includes a cross-modal attention approach that allows fusion of such outputs into context-aware classification of mental health conditions. The model was then trained and evaluated on a dataset of labeled Reddit posts preprocessed using text preprocessing, scaling of temporal features, and encoding of labels. Experimental results indicate that the proposed architecture performs better compared to traditional models with a validation accuracy of 74.55% and F1-Score of 0.7376. This study presents the importance of multi-modal learning for mental health detection and provides a baseline for further improvements by using more advanced attention mechanisms and other data modalities.**

## I. Introduction

Mental health problems such as depression, anxiety, bipolar disorder, and schizophrenia are major worldwide health challenges, impacting over a billion people each year, or more than 16% of the global population [1]. These disorders have a significant influence on cognitive, emotional, and social elements of life, with long-term implications for society and the economy [2]. Despite their prevalence, mental health issues go undiagnosed and untreated in many parts of the world due to cultural stigma, a lack of access to professional care, and a lack of efficient diagnostic tools [3]. Traditional diagnostic techniques rely on self-reporting and professional examinations, which, while valuable, are frequently reactive rather than proactive. This delay in diagnosis and treatment increases the burden on both affected individuals and healthcare systems [4][5].

During last few years, social media platforms, such as Reddit, Twitter, and Facebook, became rich sources of user-generated texts that provide useful insight into the psychological state and behavior of active users [6][7]. Social media allows users to publicly share their experiences, troubles, and feelings, introducing an unparalleled opportunity to monitor and study mental health indicators in real time. Previous research has found that linguistic patterns, attitude, and social media posting activities are strongly associated with mental health disorders [8][9]. Individuals suffering from depression, for example, frequently express themselves negatively, use self-centered language, and engage in specific temporal posting patterns [10].

While the potential of social media for mental health monitoring is obvious, current computational approaches for processing this data have numerous drawbacks. Traditional machine learning algorithms, like support vector machines and random forests, rely most on handmade characteristics and do not reflect the quality and context of language [11]. More recently, deep learning models such as CNNs and RNNs have been applied to overcome these difficulties by learning features directly from data [12]. While CNNs are good at capturing local language information, LSTMs do a better job of modeling sequential dependencies. However, these methods are often constrained by their inability to process both forward and backward contextual information simultaneously, limiting their effectiveness in understanding the complex patterns associated with mental health conditions [13][14].

Bidirectional long short-term memory (BiLSTM) networks overcome these restrictions by processing input sequences both forward and backward, allowing the model to collect more contextual information [15]. BiLSTM models have shown some promise in mental health classification, particularly when combined with enhanced embeddings such as word2vec or FastText, which enrich the representation of linguistic data [16][17]. Prior work has mostly relied solely on textual data, without taking into consideration temporal aspects that characterize social media activity, like time of day, day of week, and posting frequency that are important to understand behavioral patterns [18].

Temporal features can provide vital information about a person's mental state. For example, inconsistent posting times or greater activity late at night could indicate heightened discomfort or insomnia, both of which are significant indications of numerous mental health issues [19]. Despite their importance, temporal patterns are often treated as secondary elements or completely neglected in many current models. The integration of temporal and linguistic elements into a coherent framework is a relatively unexplored field of research [20].

In light of these pitfalls, the study proposes a multi-modal deep learning system for the early diagnosis of mental health disorders based on social media data. The proposed model uses a BiLSTM network to analyze textual and temporal features, capturing characteristics of language and behavior. There is also a cross-modal attention mechanism in the architecture that allows it to dynamically weigh textual and temporal data more or less according to their relative importance in the prediction. This method ensures that the model can adaptively focus on the most informative features, which boosts classification accuracy and generalization [21].

The model is tested on a dataset of Reddit posts that has rich textual content and timestamp information. The preprocessing processes included cleaning the text data, encoding the subreddit labels, and scaling temporal features to achieve numerical stability. The results show that the proposed model outperforms traditional deep learning algorithms, with validation accuracy of 74.55% and validation loss of 0.7130. These results evidence that multi-modal learning captures the complex interaction of language and temporal patterns related to poor mental health conditions [22].

This study contributes to the growing body of work in computational mental health with a solid and extensible framework that capitalizes on the complementary strengths of linguistic and temporal analysis. It also underlines the importance of incorporating behavioral data into predictive models, thus laying the ground for future progress in this topic.

The remainder of this paper is organized as follows. The next section presents the related work done in the past. Section III presents the motivation and our contribution. Section IV shows from where we took the dataset, what we explored from dataset and how we processed it, Section V present our proposed model, Section VI shows experiment setting and evaluation and Section VII draws conclusions.

## II. Related Work

The subject of mental health detection using social media data has advanced significantly in recent years, with a focus on applying computer tools to assess user-generated content. Machine learning techniques including support vector machines (SVMs), Naïve Bayes (NB), and decision trees have been used to categorize mental health statuses. These models are primarily reliant on handcrafted features, which limits their capacity to detect complicated and nuanced patterns in data [21][23].

Deep learning algorithms have emerged as powerful alternatives, with improved performance achieved by learning features directly from raw data. Convolutional neural networks (CNNs) are commonly used to extract local language aspects from textual data, but recurrent neural networks (RNNs), particularly long short-term memory (LSTM) networks, excel at modelling sequential dependencies [6][15]. However, LSTM models can only process data in one direction, which limits their ability to adequately capture contextual information.

BiLSTM networks overcome this by processing input sequences in both forward and backward orientations, thus enabling the model to capture more contextual information. Trotzek et al. (2020) proved the usefulness of BiLSTMs in early identification, emphasizing their ability to model linguistic and temporal connections [5]. Further refinements have been made to this strategy by combining BiLSTM and BERT embeddings, which results in large gains in mental health classification tasks [7].

Recently, multi-modal approaches that incorporate language and temporal features have become popular for mental health detection. These approaches leverage complementary data modalities to provide a comprehensive view of the mental health states. Research has been done which investigated the integration of linguistic and behavioral aspects for mental health classification, underlining the value of merging diverse data sources [19].

Attention mechanisms have enhanced performance in deep learning models by enabling a dynamic weighting mechanism that could estimate the relative importance of each feature concerning the prediction goal. Although first proposed as an attention method for machine translation, today, from its first uses, its application has reached a variety of fields such as detecting mental health status [13]. Cross-modal attention mechanisms that combine features coming from several modalities proved very successful in balancing the contribution of linguistic and temporal features within multi-modal frameworks [24].

The need to evaluate computational models using robust metrics including precision, recall, and F1-score has already been highlighted in existing research [17].

Apart from language and temporal analysis, advanced embedding techniques were used to enhance the representation of features. FastText and GloVe embeddings are usually utilized to represent words in a dense vector form that expresses semantic and syntactic relations between words [12]. These embeddings have been of great help in resolving issues related to out-of-vocabulary and enhancing the generalization capability of deep learning models [9].

This paper extends prior work by presenting a multi-modal framework that incorporates BiLSTM networks for the analysis of linguistic and temporal features, augmented by a cross-modal attention mechanism. The methodology overcomes some limitations of single-modality methods and shows the potential of multi-modal learning in improving early diagnosis related to mental health disorders.

## III. Motivation and Contribution

This study addresses the limitations highlighted in the introduction and related work, presenting a novel multi-modal deep learning framework for early detection of mental health conditions. The proposed model focuses on textual and temporal features with Natural Language Processing methods using BiLSTM and LSTM networks, respectively, and then utilizes a cross-modal attention mechanism to weight each modality. By doing so, the model selects only the most informative features to improve the accuracy and generalization of classification. Temporal features extracted from user-generated data supplement the linguistic analysis and capture vital behavioral patterns in the assessment of mental health.The major contributions of this work are highlighted below:

a) This work is the first to classify mental health using both textual and temporal characteristics, addressing a previously unexplored area. This model uses BiLSTM for text and LSTM for temporal analysis to capture linguistic and behavioral trends.

b) The model uses a novel attention mechanism to prioritize textual or temporal information based on their relevance to the classification job.

Extensive testing was performed on a dataset of Reddit posts using the model, which outperforms standard models such as BiLSTM, LSTM, and CNN, as well as simply textual deep learning models, with an accuracy of 74.55% and a loss of 0.7130 during validation. This study lays the groundwork for additional research into multi-modal approaches to mental health assessment, highlighting the need of combining behavioral and linguistic data to increase prediction accuracy.

## IV. DATASET

The dataset utilized in this study focuses on mental health and was sourced from Kaggle, specifically from the repository titled "Mental Disorders Identification Reddit NLP" (accessible at [Kaggle Dataset] [25]. It contains user-generated text, which belongs to any one of the several types mentioned, including posts and commentary about anxiety, depression, and BPD, each representing a particular type in the subreddit r/mental health. The dataset has a combined total of 701,809 data points.

Each entry in this dataset has the following columns:

- **title**: The heading or title of the Reddit post.
- **selftext**: The body or main content of the post.
- **created_utc**: The timestamp indicating when the post was created.
- **over_18**: A label indicating whether the post contains Not Safe For Work (NSFW) content.
- **subreddit**: The subreddit on which the post was published, serving as the class label for this study.

Table 1 represents a sample of what the dataset structure looks like to give an idea of the kind of information available for analysis. The data were further organized into an Excel spreadsheet for preprocessing and categorization. The mental health categories included in this study span seven unique labels: borderline personality disorder, anxiety, depression, bipolar disorder, mental illness, and schizophrenia.

This forms a rich dataset that can build a better understanding of linguistic and behavioral patterns related to mental health conditions and hence is suitable for the training and evaluation of the proposed multimodal deep learning framework. The text-based and metadata features allow for a comprehensive analysis of indicators related to mental health.

TABLE I. MENTAL HEALTH REDDIT DATASET EXAMPLE

| **Title** | **Selftext** | **Created_Utc** | **Over_18** | **Subreddit** |
|---|---|---|---|---|
| Why do I see self harm as enjoyable (I did stop due to increased surveillance but this question has been on my mind recently) | Well to put it simply, I recently started self-harming enough to leave scars on my body. Before I was kinda hesitant to feel pain and even when I held a knife so badly wanting to cut, I never could. But now I can cut myself pretty easily and started poking my arms with a needle very easily. Like I'm somewhat proud of my scars. But only slightly ashamed when my professors literally calling me out in front of the entire class about why I have bandages on (so I make up stupid lies). But other than that I kinda flaunt them to people around me. Like yeah I'm mentally unstable what about it | 1668668048 | TRUE | mental illness |

### A. Data Exploration

The dataset utilized in this study provides source of information for understanding patterns in mental health-related discussions on Reddit. During the data exploration phase, several key observations and transformations were made to ensure the dataset was suitable for analysis and classification.

The dataset was categorized into key mental health subreddits that act as labels for classification. These subreddits include Borderline Personality Disorder (BPD), Anxiety, Depression, Bipolar Disorder, Mental Illness, Schizophrenia.

A preliminary analysis of the information showed that 2.7 percent was marked NSFW, reflecting Figure 1. Most of the content about NSFW posts had their contents removed and are, therefore, worthless in this study. Hence, all those rows have been dropped to persist only with quality data relevant to the studies.

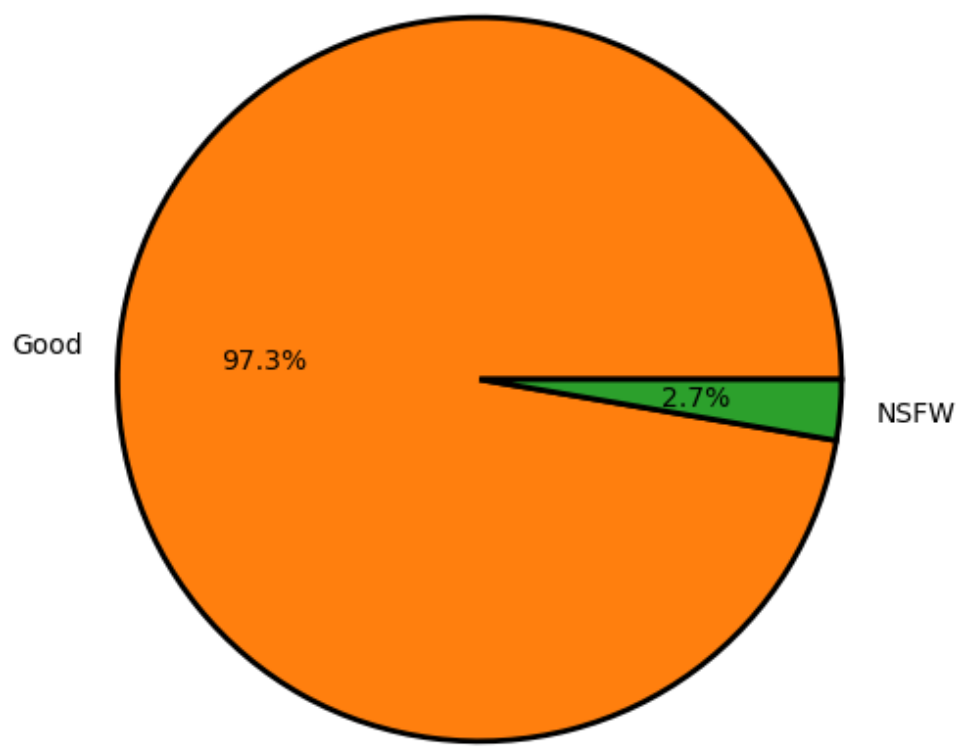

Fig. 1. Distribution of Posts by Age group

The dataset ranges from posts in 2012 until 2022. Therefore, the number of postings that denote despair and anxiety gradually increased, especially in post COVID period 2020. Figure 2 shows how fast growth turned into an upward pattern aggressively in the same year, 2022. To make sure all dates have an equal number of posts, the dataset was restricted to range from 2021 up to 2022 only.

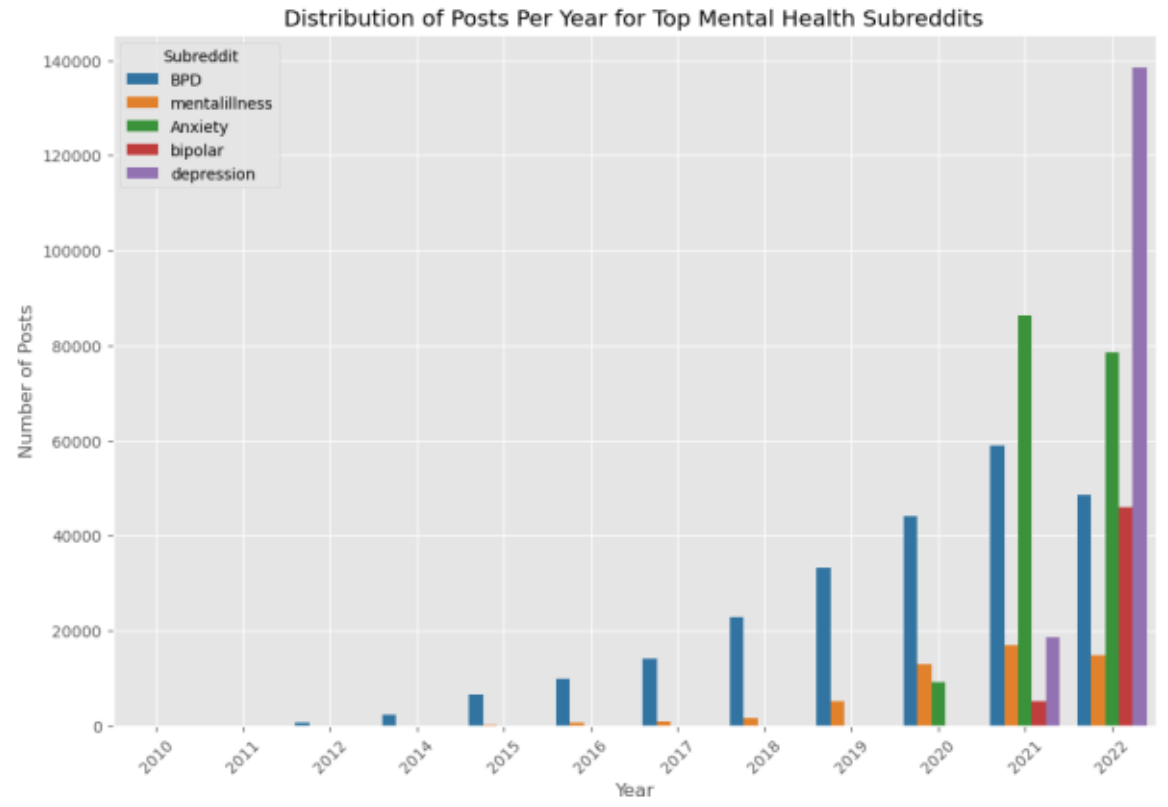

Fig. 2. Distribution of Posts Per Year for Mental health Subreddits

As shown in Figure 3 a word cloud display was made to identify key words across the dataset in highest use. The following visual focuses on dominating phrases in an effort to demonstrate recurring themes and topics of the postings covered. The prominent leading words, such as "help," "feel," "like," or specific terms with their mental health disorders, indicate that users clearly want to seek help or talk about their experiences in particular ways.

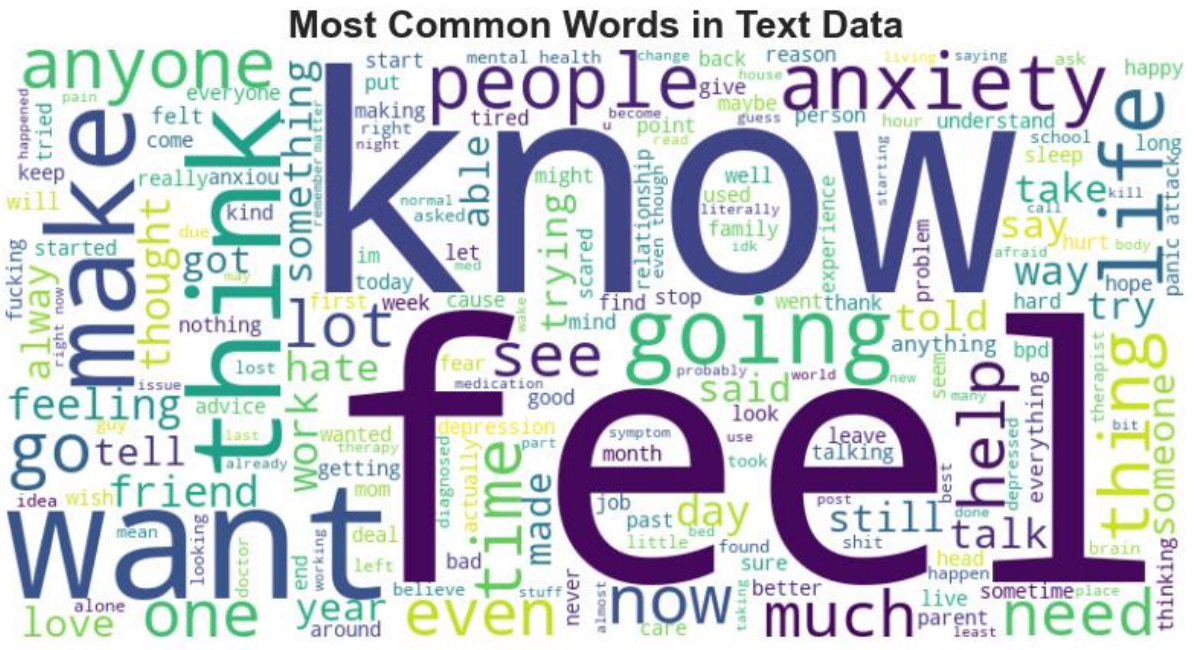

Fig. 3. Word Cloud for Most Common Words

The posting timing analysis, represented in Figure 4, has a very clear trend—most posts are between 4 PM and 6 AM. This might indicate either the timing of emotional suffering or users' willingness to share experiences during less organized times of the day. This temporal pattern gives important insights into user behavior and can be used to plan timely interventions for those in need.

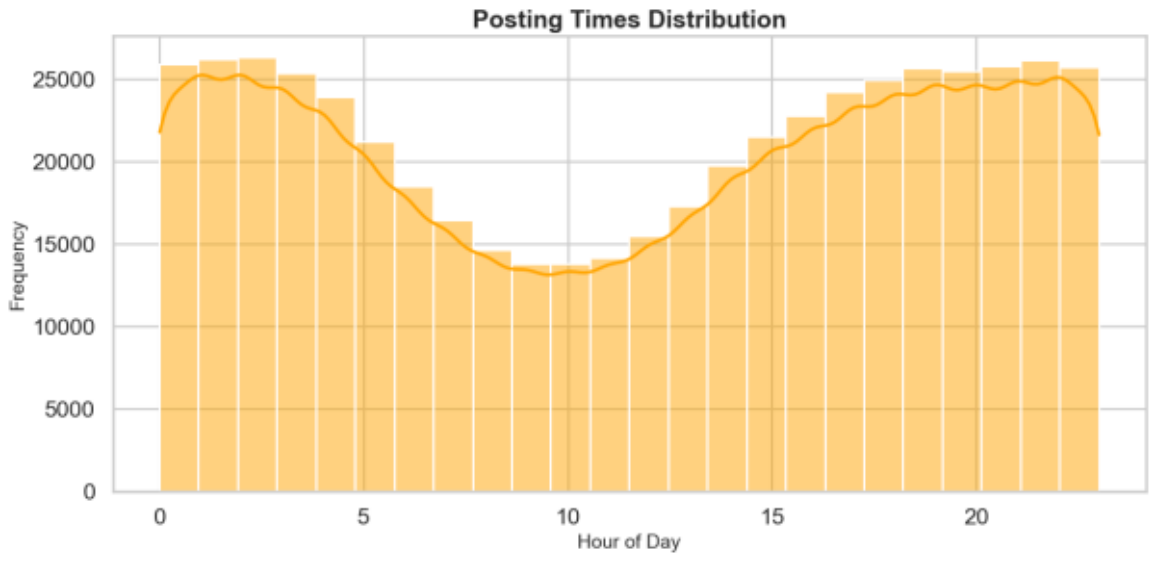

Fig. 4. Word Cloud for Most Common Words

The length of posts in different mental health subreddits was also analyzed to show the pattern of communication. Figure 5 shows that the posting in subreddits like BPD and Mental Illness are way longer than in categories like Schizophrenia. In subreddits like BPD and Mental Illness, people often make very long posts describing their disorder and symptoms in detail, whereas posts in other categories are short, mainly to seek quick advice or short questions.

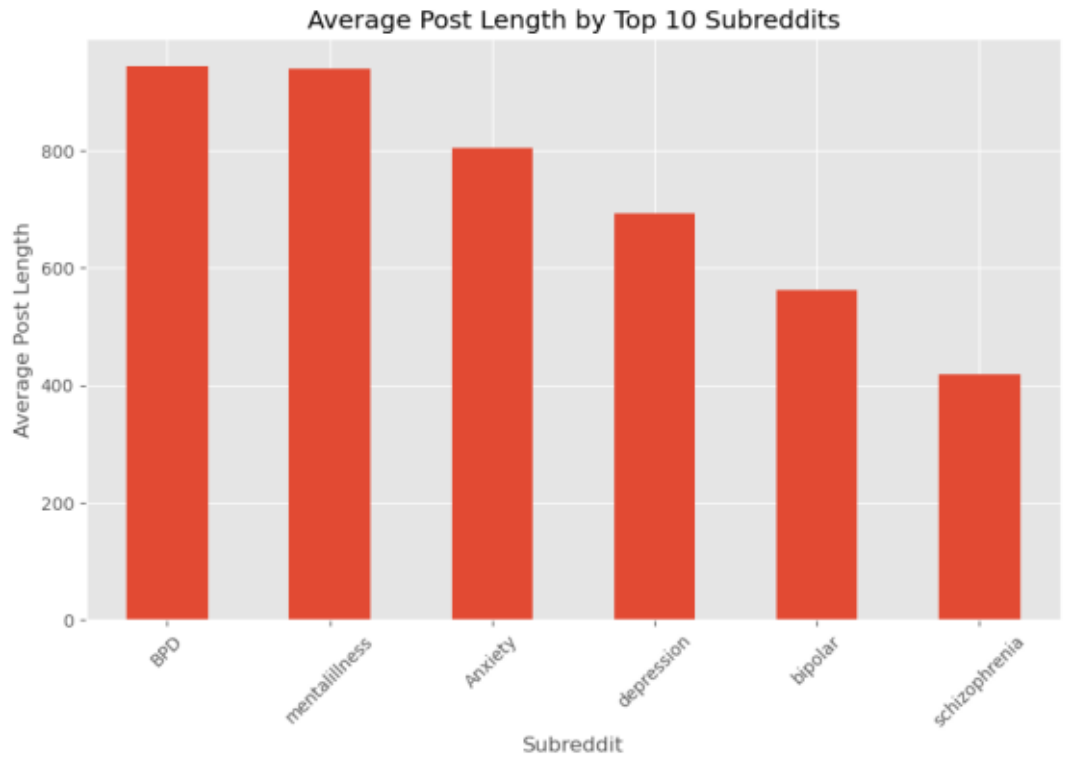

Fig. 5. Subreddits Post Lengths

The result of these exploratory analyses underlined some key features in the dataset that weighed, consequently, on the preprocessing and feature engineering decisions. This analysis brought direct influence to model construction, underlining the necessity of combining linguistic, temporal, and behavioral patterns into an accurate mental health classifier.

## *B. Data Preprocessing*

To prepare the dataset for analysis, we implemented a thorough data preprocessing pipeline to handle both textual and temporal aspects of the data.

For textual data processing, combined the "title" and "selftext" fields of every post into one column called "combined_text" to put all the textual information into one feature. The second step in data preparation involves ensuring that the title is combined with the actual content of the post while taking into consideration not to exclude or lose important information across titles or body content. The filtering-out process is applied against any null values across either columns—"title" or "selftext"—to prepare this dataset, keeping records intact and issues at bay.

As shown in Figure 6, the text data are cleaned of unnecessary punctuation, stop words, and special characters because these usually add noise and do not contribute toward meaningful patterns in natural language processing tasks. Further betterment in the quality of text representation was achieved with the application of Word2Vec embeddings, which transform words into dense numerical vectors while capturing semantic relationships among them. Further lemmatization was performed in order to reduce the words into base or root forms of the word to make different forms of words to a common representation. Tokenization is done to split combined text into individual words so that finer resolution and representation of textual data can be done.

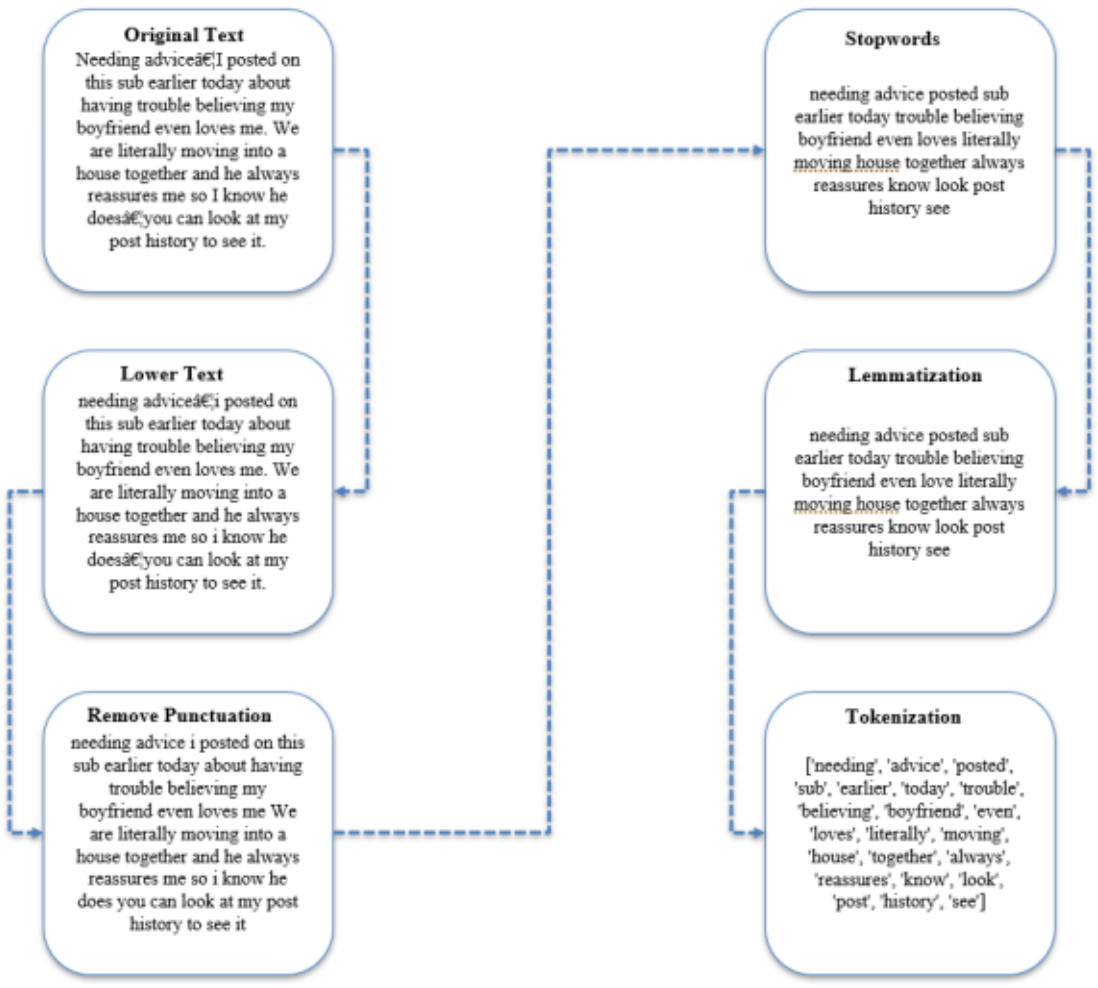

Fig. 6. Textual Data Preprocessing

For temporal feature engineering part, we derived several key features from the given UTC timestamps present in the dataset. The features that were then extracted from each observation included "month", "day", "hour", and "weekday", with two boolean indicators: "is_working_hour" and "is_weekend". The feature "is_working_hour" was for the capturing of whether a post had been made during typical work hours, and "is_weekend" showed if the post was published on a weekend. These temporal features were of great importance in understanding the posting behaviors—for example, whether the users shared their experiences during working hours or late at night.

All these features should be prepared for being fed into machine learning models by scaling them within the range of [0, 1] using MinMaxScaler. This normalization step increased numerical stability and ensured that the time features would not dominate during model training.

The preprocessing step above has been designed to derive most of the insight from this dataset in a way that the textual and temporal features are cleaned and consistent to be taken into consideration by a multimodal deep learning model. This approach has not only improved the model in terms of better representation of the dataset but also enhances multimodal pattern capturing in terms of different data modalities using the model.

## V. Proposed Deep Learning Model

The aim of this work is to propose a multi-modal deep-learning architecture that considers both text and temporal features toward the early recognition of mental health disorders. In fact, the architecture of such a design would embed aptly complex interactions of linguistic and behavioral features characterizing users' posts on social media platforms. As shown in Figure 7 model contains a Textual Analysis Module, Temporal Analysis Module, Cross-Modal Attention, and a Classifier. Each of them is explained with relevant mathematical formulation in the section below.

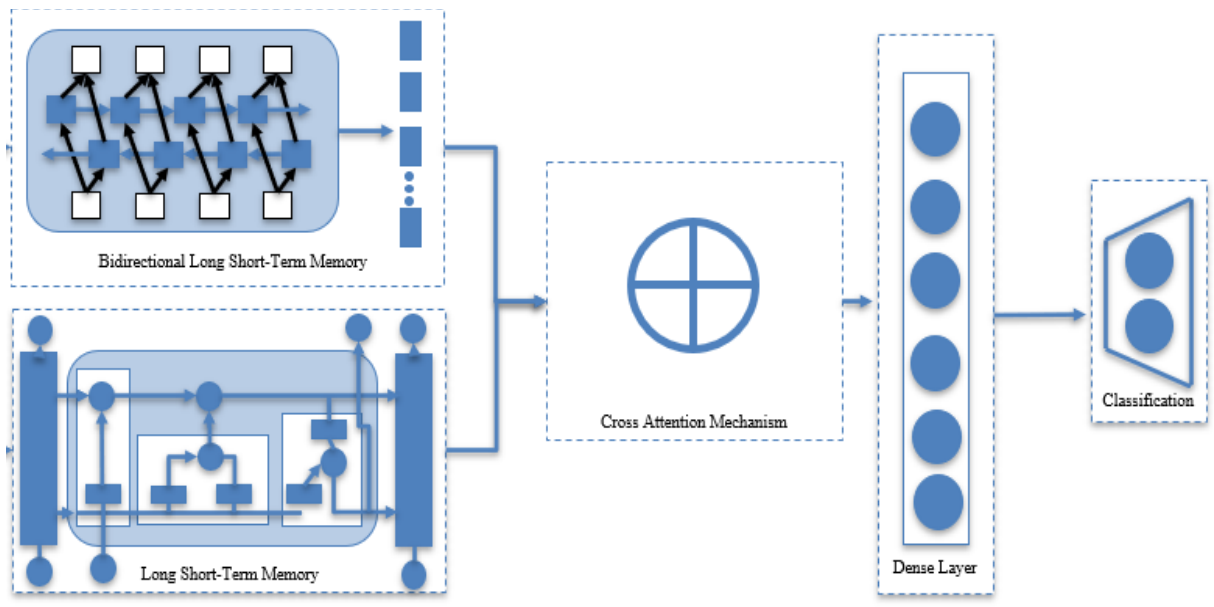

Fig. 7. Proposed Deep Learning Model

### A. *Bidirectional Long Sthort term Memory*

Bi-LSTM gives a better understanding of structural and semantic insights in a language as compared to uni-directional ones. “combined_text” feature is fed into the network for analysis using the Bi-directional Long Short-Term Memory Network, which captures information context depending on both past and future directions of the sequences. As shown in Figure 8 Bi-LSTM consists of two LSTMs:

*1) Forward LSTM:* Processes the input sequence from the first word to the last.

*2) Backward LSTM:* Processes the input sequence from the last word to the first.

Given an input sequence of word embeddings $x = [x_1, x_2, \dots, x_T]$, where $T$ is the sequence length, the forward and backward LSTMs compute the hidden states as:

$$\vec{h}_t = LSTM(x_t, \vec{h}_{t-1}) \quad (1)$$

$$\overleftarrow{h}_t = LSTM(x_t, \overleftarrow{h}_{t+1}) \quad (2)$$

The hidden states at each time step are concatenated to produce the Bi-LSTM output:

$$h_t = [\vec{h}_t; \overleftarrow{h}_t] \quad (3)$$

The Bi-LSTM outputs a sequence of hidden states $H = [h_1, h_2, \dots, h_T]$, which representing the contextual embedding for each word in the sequence. To summarize the information at sequence-level, a max-pooling operation is applied:

$$z_{text} = \max(H) \quad (4)$$

This $z_{text}$ serves as the textual embedding for further processing, capturing both sequential and contextual dependencies.

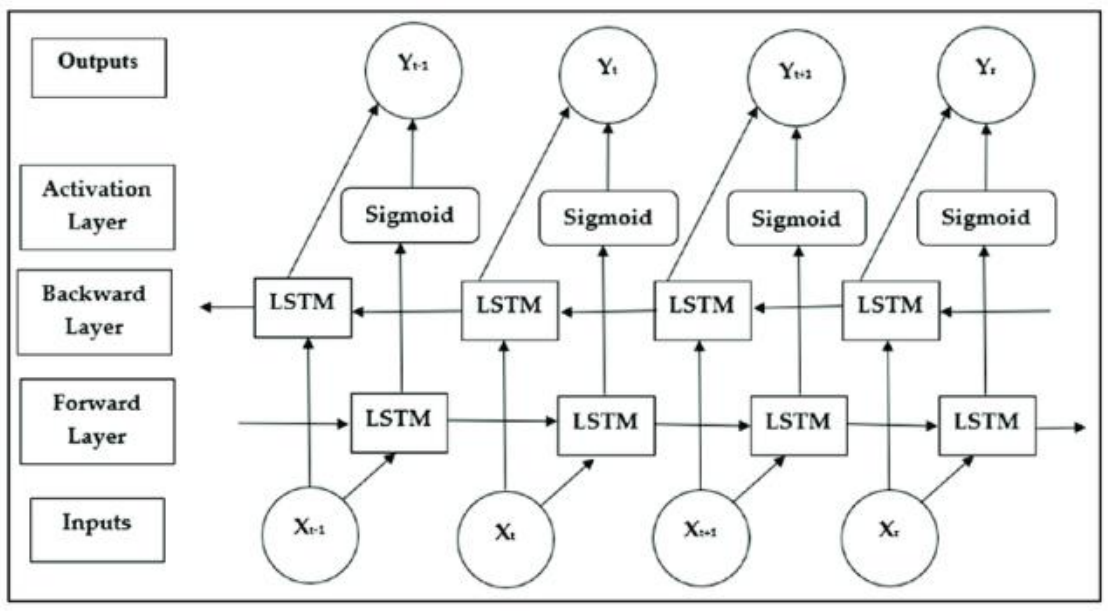

Fig. 8. Bidirectional Long Short Term Memory Network

### B. *Long Short-Term Memory*

The LSTM network analyzes the temporal features, which include "month, day, hour, weekday, is_working_hour, and is_weekend". These features capture temporal patterns in user behavior, such as time and frequency of posts that are often linked to mental health conditions. An overview of the architecture for LSTM is presented in Figure 9.

For a sequence of temporal features $t = [t_1, t_2, \dots, t_n]$, where $n$ is the sequence length, the LSTM computes the hidden states as follows:

$$h_t = LSTM(t_t, h_{t-1}) \quad (5)$$

The final hidden state $h_n$ represents the temporal embedding $z_{time}$ summarizing the sequential dependencies in the temporal data.

The LSTM’s key advantage lies in its ability to retain information over long sequences while avoiding issues like vanishing gradients, which are common in traditional RNNs. The gating mechanisms within the LSTM are defined as:

$$f_t = \sigma\left(W_f . [h_{t-1}, t_t] + b_f\right) \, (Forget\ Gate) \quad (6)$$

$$i_t = \sigma(Wi . [h_{t-1}, t_t] + b_i) \, (Input\ Gate) \quad (7)$$

$$o_t = \sigma(W_o . [h_{t-1}, t_t] + b_o) \, (Output\ Gate) \quad (8)$$

$$c_t = f_t \odot c_{t-1} + i_t \odot \tanh\left(W_c . [h_{t-1}, t_t] + b_c\right) \, (Cell\ State\ Update) \quad (9)$$

$$h_t = o_t \odot \tanh(c_t) \quad (10)$$

Here in the above-mentioned equations σ represents the sigmoid activation function, $\odot$ denotes element-wise multiplication, and W and b are learnable weights and biases.

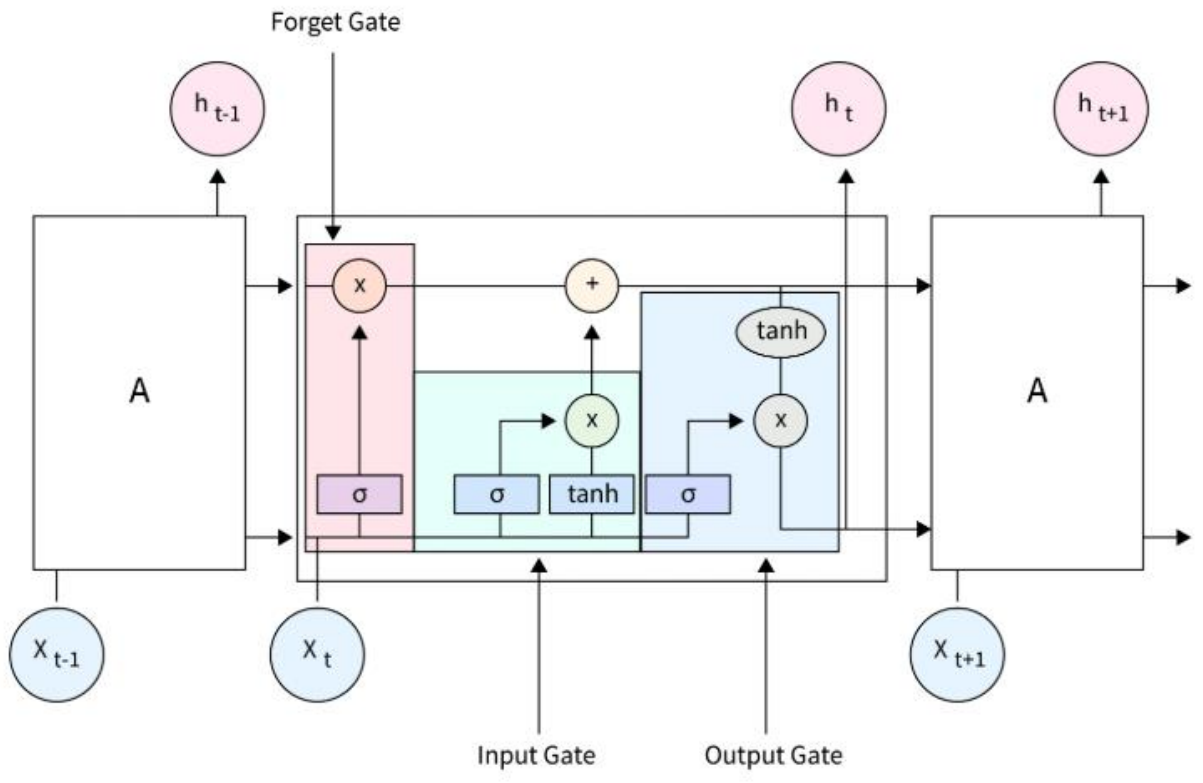

Fig. 9. Long Short Term Memory Network

## C. Cross-Modal Attention Mechanism

The cross-modal attention mechanism fuses the textual embedding $z_{text}$ and the temporal embedding $z_{time}$ by dynamically assigning weights to each modality based on their relevance to the prediction task.

The attention mechanism computes scores for both modalities as shown below:

$$\alpha_{text} = \frac{a}{a+b} \quad (11)$$

$$\alpha_{time} = \frac{b}{a+b} \quad (12)$$

Where $a = \exp(w^T \tanh(W_{text} z_{text} + b_{text}))$ and $b = \exp(w^T \tanh(W_{time} z_{time} + b_{time}))$

The fused embedding $z_{fused}$ is computed as:

$$z_{fused} = \alpha_{text} z_{text} + \alpha_{time} z_{time}$$

This mechanism uses Cross Attention to compute the attention as score as shown in Figure 10. makes sure that the model adapts to the most informative modality, depending on the context of each prediction.

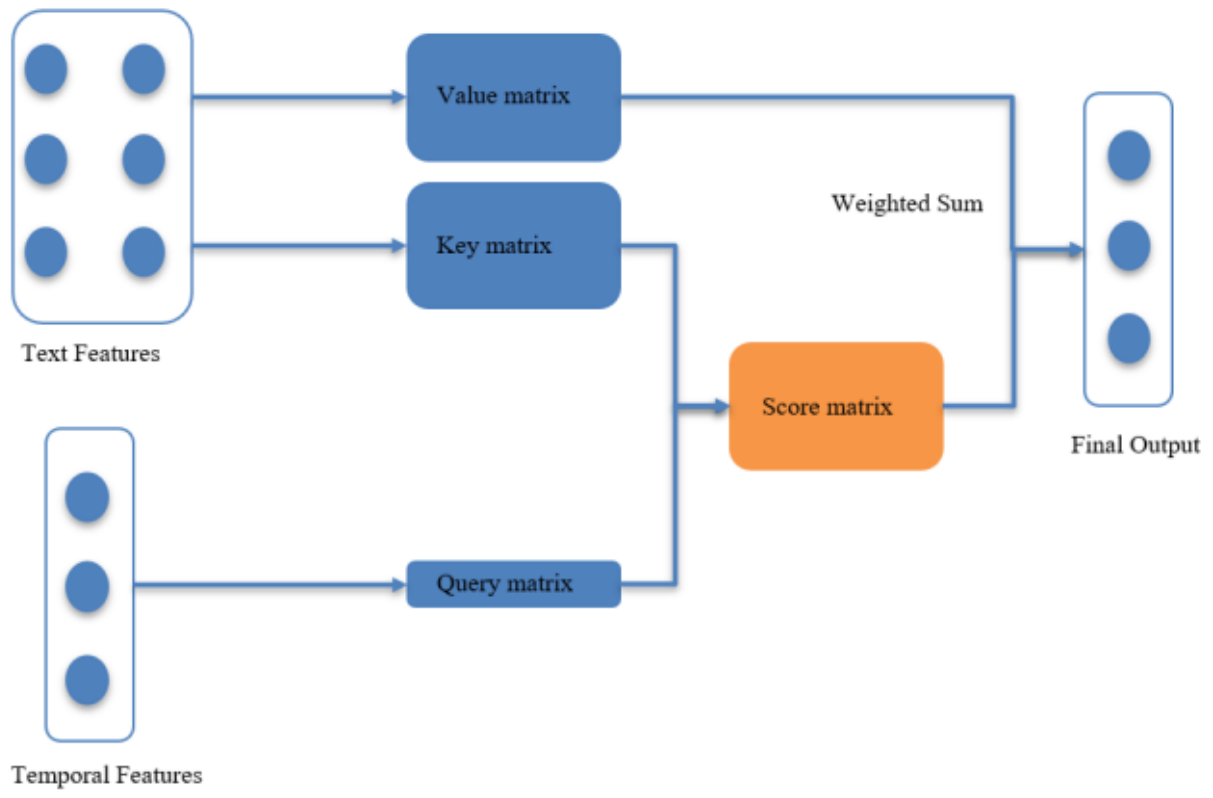

Fig. 10. Cross Attention

## D. Dense Classifier

The fused embedding $z_{fused}$ is passed through a fully connected dense layer to map the learned features to the output classes. The class probabilities are computed using the softmax activation function:

$$\hat{y} = softmax\left(W_{output} z_{fused} + b_{ouput}\right) \quad (13)$$

Here $W_{output}$ is the weight matrix of the dense layer, $b_{ouput}$ is the bias vector, and represents the predicted probabilities for each class, where C is the total number of classes.

The softmax function ensures that the output probabilities sum to 1, making it suitable for multi-class classification tasks. The probability for a given class iii is computed as:

$$\hat{y} = \frac{\exp(z_i)}{\sum_{j=1}^{C} \exp(z_j)} \quad (14)$$

Here $z_i = W_{output[i,:]} . z_{fused} + b_{output}[i]$ is the raw score (logit) for class $i$.

As shown in Figure 11 the dense classifier allows the model to interpret the high-dimensional fused feature vector $z_{fused}$ and map it to a specific mental health condition, such as depression, anxiety, or borderline personality disorder.

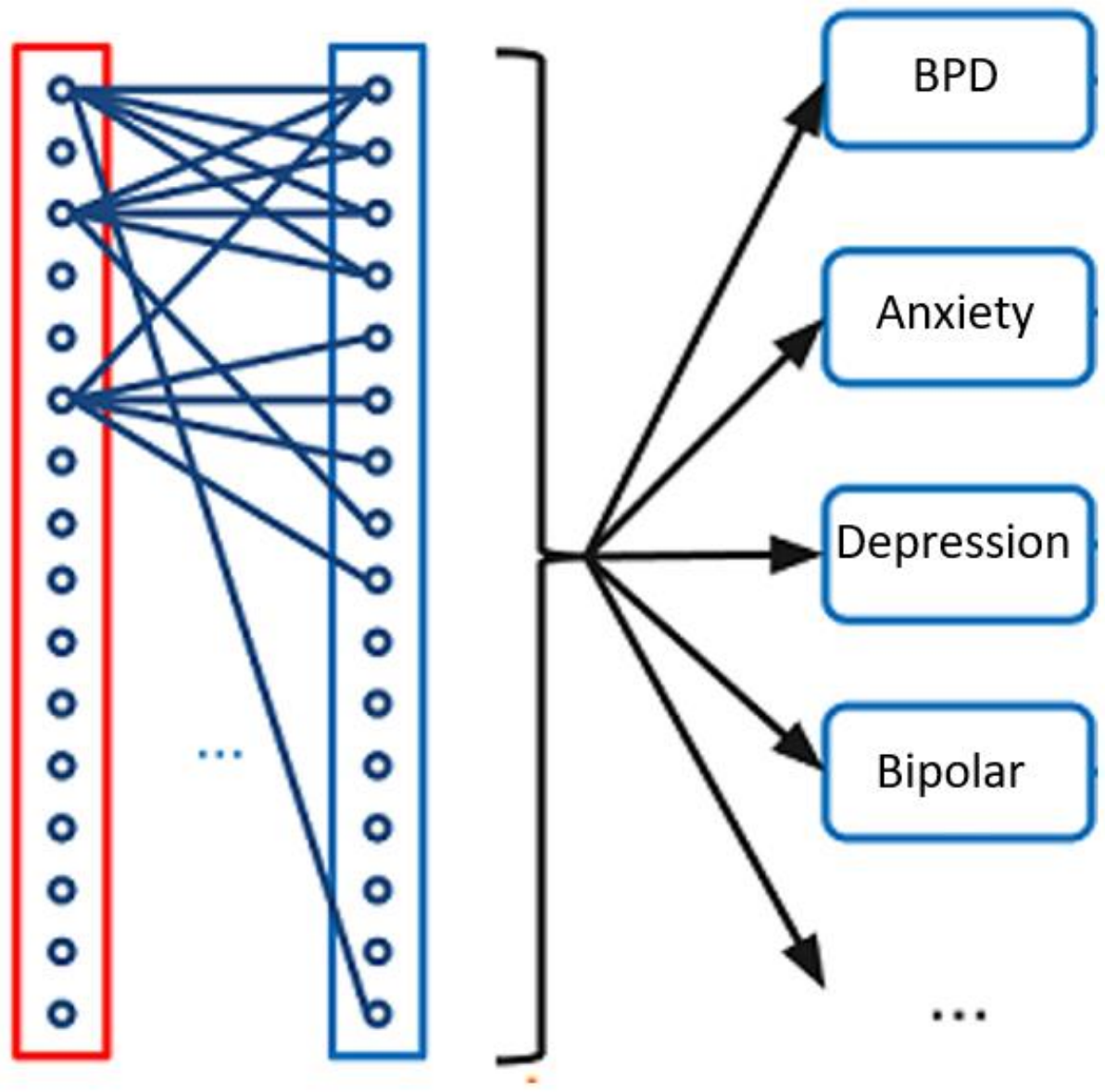

Fig. 11. Dense Classifer

# VI. Experiment and Evaluation

## A. Optimizer and Loss Function

The model is trained with the RMSprop optimization algorithm, a gradient-based optimization method that automatically adapts the learning rate for each parameter based on the magnitude of recent gradients. RMSprop works very well when training deep learning models where the objective is non-stationary, avoiding the problem of exploding or vanishing gradients.

The RMSprop optimizer updates the parameters as follows:

$$g_t = \nabla L(\theta_t) \quad (15)$$

$$E[g^2]_t = \beta E[g^2]_{t-1} + (1-\beta) g_t^2 \quad (16)$$

$$\theta_{t+1} = \theta_t - \frac{\eta}{\sqrt{E[g^2]_t + \varepsilon}} g_t \quad (17)$$

Here $\theta_t$ is the model parameters at time step $t$, $g_t$ is the Gradient of the loss $L$ with respect to $\theta_t$, $E[g^2]_t$ is the exponentially weighted moving average of squared gradients, $\beta$ is the smoothing factor for the moving average, $\eta$ is the learning rate which controls the step size during optimization and $\varepsilon$ is the small constant added for numerical stability.

RMSprop ensures the parameters with large gradients are scaled down while those with small gradients are scaled up,

thereby making efficient training across diverse features and reducing the oscillations in the process of optimization.

The optimization is done by minimizing the categorical cross-entropy loss function:

$$L = -\sum_{i=1}^{C} y_i \log(\hat{y}_i) \quad (18)$$

Where $y_i$ is the true label (one-hot encoded) and $\hat{y}_i$ is the predicted probability for class *i*.

This combination of the RMSprop optimizer with the categorical cross-entropy loss function makes the model learn effectively to distinguish classes of mental health while keeping the convergence smooth and stable during training period.

### B. Evaluation Metrics

To assess the performance of the proposed model, we utilized four widely accepted evaluation metrics: Precision, Recall, F1-Score, and Accuracy. These metrics provide a broad view of the model's ability to correctly classify mental health conditions while minimizing false predictions.

Precision is the ratio between the true positive predictions with respect to all instances predicted as positive. It gives an idea of how well a model will perform in avoiding false positives in each class. Precision is important when one wants to reduce false positives, such as classifying posts that are not about mental health as being about mental health. The formula of precision is given below:

$$Precision = \frac{True\ Positive\ (TP)}{True\ Positive\ (TP) + False\ Positive\ (FP)} \quad (19)$$

Recall, also known as sensitivity or true positive rate, basically measures the proportion of actual positive instances among all the true positive predictions. It reflects a model's ability to detect all instances of a class. Recall ensures that most relevant cases are captured by the model, which is extremely important in identifying at-risk cases within a mental health dataset. The formula of recall is given below:

$$Recall = \frac{True\ Positive\ (TP)}{True\ Positive\ (TP) + False\ Negative\ (FN)} \quad (20)$$

The F1-Score is the harmonic mean of precision and recall. It provides a balanced measure considering both false positives and false negatives; hence, it's useful on imbalanced datasets. F1 Score can be viewed as the measure that balances precision and recall, yielding one score for assessing the global performance of the model. The mathematical representation of F1 Score is stated by the formula below:

$$F1 - Score = 2.\frac{Precision * Recall}{Precision + Recall} \quad (21)$$

Accuracy is the measure of the right set of instances classified vis-a-vis the total number present in the dataset. Thus, it gives an overarching look at how well, globally, the model is really doing across all classes. While accuracy provides the general idea about the model's classification capability, it is supplemented by the other metrics to reach an in-depth analysis. Accuracy can be calculated as:

$$Accuracy = \frac{True\ Positive\ (TP) + True\ Negative\ (TN)}{Total\ number\ of\ Instances} \quad (22)$$

These evaluation metrics altogether provide a comprehensive assessment of the performance of the model that includes not only the ability to predict precisely but also the effectiveness of finding the relevant cases for different mental health categories.

### C. Hyperparameter Setting

The proposed model is trained and evaluated with a carefully chosen set of hyperparameters to ensure the model has learned optimally and is doing the best classification. For the text data, the vocabulary size is limited to the 10,000 most frequent tokens; each token is represented by a 128-dimensional word embedding that captures semantic and syntactic relations among words, providing rich input for the BiLSTM layer. Among others, the BiLSTM architecture utilized 64 hidden units to learn long-range sequential dependencies and represent contextual information from both forward and backward directions of text. On similar lines, for temporal data, a layer of LSTM used with 64 hidden units served the purpose of modeling a range of behavioral patterns reflecting the posting times and posting frequencies—a typical set of critical indicators towards early diagnosis.

The dropout rate of 0.6 was used to prevent overfitting, ensuring generalization by randomly dropping a fraction rate of neurons during training. For the RMSprop, the learning rate was set to 0.0005, smoothing factor α to 0.99 to balance gradients over time, momentum to 0.9 to accelerate convergence, and epsilon to $10^{-8}$ to avoid division by zero. Other measures to prevent overfitting included the L2 regularizer with a weight decay of $10^{-5}$ .

The model would be trained through 30 epochs, using a batch size of 64, leveraging the categorical cross-entropy function since it deals with a multi-class problem. Text sequences that are either too few or too many items were preprocessed by padding/truncating every sequence into 100 tokens, thus facilitating a standard dimensionality of feature vectors for inputs across the dataset. Stratification was performed to preserve class distribution in both training and evaluation. Therefore, stratified sampling of the data into train-test splits was done.

The best performance model was selected based on the F1-Score on the validation set during training since the F1-Score is a balanced measure of Precision and Recall. These hyperparameter settings and training strategies allowed the model to realize good predictive accuracy without any compromise on either computational efficiency or robustness.

### D. Results

Various relevant metrics, such as validation loss and accuracy, precision, recall, and F1-Score, have been used to evaluate the performance of the proposed model. Results from the model evaluation show that the model is effective with a high performance in categorizing mental health disorders using multimodal inputs. Besides, the model reported 0.7130 as the minimum validation loss and 74.55% as the validation accuracy, which indicates good generalization on previously seen data. The model performances are comprehensively outlined in Table 2, having a precision of 0.7407, a recall of 0.7455, and an F1-score of 0.7376. These three balanced metrics testify to the performance of the model that is regular for most of the various evaluation criteria while being very effective in minimizing both false positives and false negatives.

TABLE II. EVALUATION RESULTS

| | Precision | Recall | Accuracy | F1-Score |
|---|---|---|---|---|
| **Our Model** | 0.74007 | 0.7455 | 0.7455 | 0.7376 |

In Figure 12 Confusion matrix shows the performance across different classes. The distribution provides true positives and misclassification across the class, so a model is capable of accurately distinguishing a class as belonging while showing patterns for these very errors. The highest falls in the classification of instances by the model in "Class 3," which represented Depression, at 26,467. On the other hand, some misclassifications were observed between the two closely related classes "Class 1" for Anxiety and "Class 2" for BPD, probably because some features are shared.

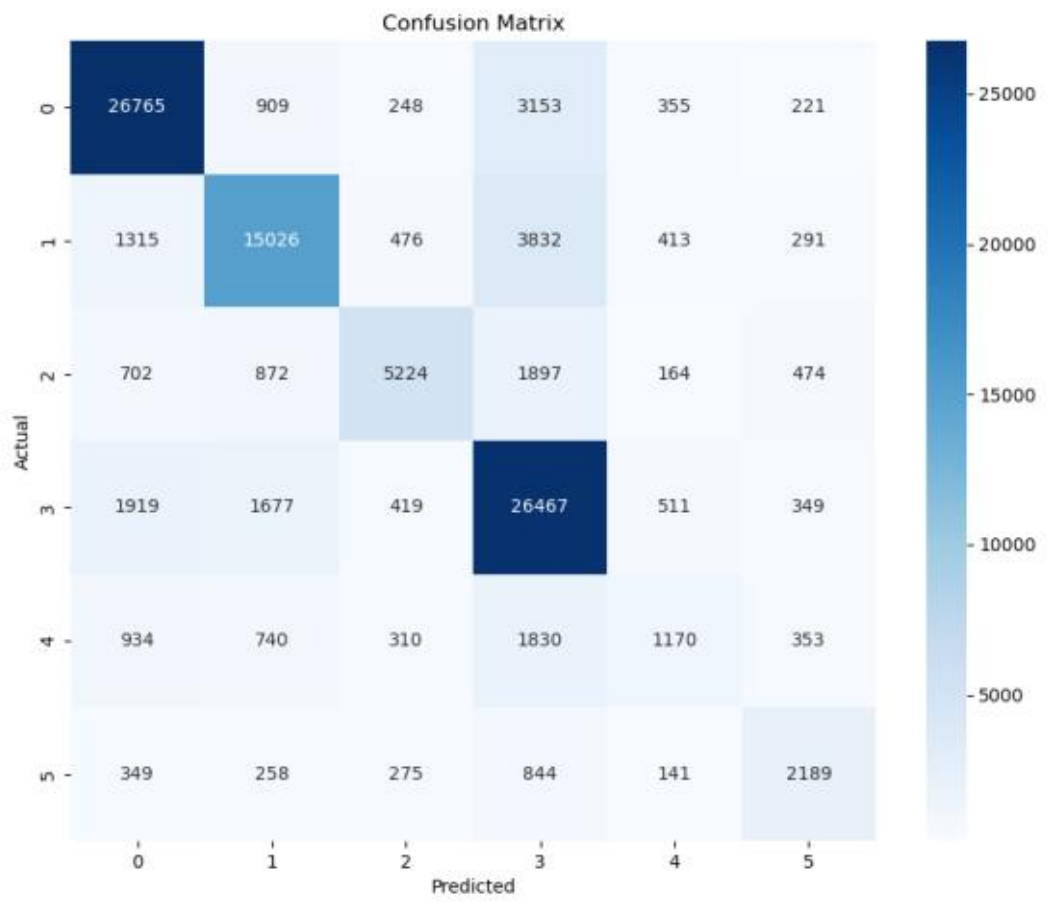

Fig. 12. Confusion Matrix

The results show the robustness of the proposed model in multi-class classification tasks for mental health prediction. This level of accuracy much owes to the incorporation of textual features and temporal features, including the attention mechanism, which greatly reduce the errors of classification. These results support the validity of this model in real-world scenarios concerning the analysis and interventions for mental health.

### *E. Comparison*

The proposed model's performance was compared to baseline architectures such as BiLSTM, LSTM, and CNN, which were trained using only text data, as they are typically utilized in previous research. In contrast, the proposed model employs a cross-modal attention mechanism to handle two modalities textual and temporal data at the same time. Table 2 compares precision, recall, and F1-Score for all models trained and evaluated under identical conditions over 30 epochs.

TABLE III. COMPARISON WITH BASELINE MODELS

| | Epochs | Precision | Recall | F1-Score |
|---|---|---|---|---|
| **Our Model** | 30 | 0.74007 | 0.7455 | 0.7376 |
| **BiLSTM** | 30 | 0.7273 | 0.7334 | 0.7222 |
| **LSTM** | 30 | 0.7171 | 0.7249 | 0.7157 |
| **CNN** | 30 | 0.7298 | 0.7179 | 0.7034 |

The findings demonstrate the benefits of combining temporal variables with textual data. With a precision of 0.7407, recall of 0.7455, and F1-Score of 0.7376, the suggested model outperforms baseline models i.e. BiLSTM, LSTM, CNN that depend purely on textual data. The addition of modality i.e. temporal data and the cross-modal attention mechanism allows the model to capture both linguistic patterns and temporal behaviors, resulting in higher classification accuracy and F1-score for this task. Among pure text-based models, the best F1-Score of 0.7222 is attained by the BiLSTM since it captures bidirectional dependencies in text. Because of its unidirectional processing, LSTM achieved somewhat worse with an F1-Score of 0.7157. The CNN, which mostly extracts out the local features, contributes the lowest F1-Score of 0.7034, unable to handle the sequentially linked nature of textual input elegantly. Our model required twice as much training time as the BiLSTM and had nearly double the number of parameters. This comparison underlines the limits of previous techniques that rely only on textual features, highlighting the superiority of the proposed framework in utilizing both textual and temporal features for mental health categorization tasks.

## VII. CONCLUSION

In this paper, we describe a new multi-modal deep learning framework for the categorization of mental health illnesses that incorporates textual and temporal information using a BiLSTM, LSTM, and a cross-modal attention mechanism. The suggested model uses user-generated postings from social media sites like Reddit and Twitter to gather language patterns as well as behavioral temporal data, allowing for a more thorough understanding of mental health disorders. By tackling the underexplored research topic of different modalities integration in this field, the proposed system outperforms traditional text-only models. Our evaluation results show that (a) adding temporal features alongside text features greatly improves classification performance, outperforming conventional models, (b) the suggested model outperformed baseline architectures that exclusively used textual data, such as BiLSTM, LSTM, and CNN, in terms of evaluation metrics i.e. precision, recall, and F1-Score, (c) the cross-modal attention mechanism dynamically prioritizes textual or temporal variables, increasing the model's predictive power, (d) The derived temporal variables, such as posting times and patterns, were crucial in comprehending behavioral trends, allowing the model to gain insights about user activity and its relationship to mental health, (e) combining textual and temporal data provides a solid foundation for multi-modal classification in the mental health sector. This study demonstrates the potential of combining several modalities, such as textual and temporal data, to improve mental health illness categorization tasks and provides a solid foundation for future research. Future study could focus on expanding the model to include new modalities, such as images or sounds, to improve its effectiveness for this task, as every percentage point increase in accuracy is significant.